# IMAGE QUALITY ASSESSMENT FOR FOLIAR DISEASE IDENTIFICATION (AGROPATH)

*Nisar Ahmed [1], Hafiz Muhammad Shahzad Asif [2], Gulshan Saleem [3]
and Muhammad Usman Younus * [4]*

[1] Ph.D. Scholar, [2] Associate Professor, Department of Computer Science and Engineering, University of Engineering and Technology, Lahore, [3] Ph.D. Scholar, Department of Computer Science, COMSATS University Islamabad, (Lahore Campus), Lahore, [4] Researcher, Ecole Doctorale Matheematiques, Informatique, Telecommunication de Toulouse, Paul Sabatier University, Toulouse, France

*Corresponding author's email: usman1644@gmail.com

## ABSTRACT

Crop diseases are a major threat to food security and their rapid identification is important to prevent yield loss. Swift identification of these diseases are difficult due to the lack of necessary infrastructure. Recent advances in computer vision and increasing penetration of smartphones have paved the way for smartphone-assisted disease identification. Most of the plant diseases leave particular artifacts on the foliar structure of the plant. This study was conducted in 2020 at Department of Computer Science and Engineering, University of Engineering and Technology, Lahore, Pakistan to check leaf-based plant disease identification. This study provided a deep neural network-based solution to foliar disease identification and incorporated image quality assessment to select the image of the required quality to perform identification and named it Agricultural Pathologist (Agro Path). The captured image by a novice photographer may contain noise, lack of structure, and blur which result in a failed or inaccurate diagnosis. Moreover, Agropath model had 99.42% accuracy for foliar disease identification. The proposed addition can be especially useful for application of foliar disease identification in the field of agriculture.

## INTRODUCTION

Crop disease identification is vital for food security as timely diagnosis will assist in taking the necessary measures to avoid crop loss. Foliar disease identification is an option as most crop diseases leave a particular footprint on the foliar structure. Moreover, wide spread penetration of smartphones has made it possible to use a smartphone-assisted diagnosis of foliar disease. Earlier research in this domain (Baranwal *et al.,* 2019; Brahimi *et al.,* 2017; Durmuş *et al.,* 2017; Hlaing and Zaw, 2017a; Islam *et al.,* 2017; Pertot *et al.,* 2012; Suryawati *et al.,* 2018; Younus *et al.,* 2016) has used small data sets, typically containing few crops and disease but extension of the crop types and coverage of broad spectrum of disease is important. One such effort is done by (Mohanty *et al.,* 2016), who have proposed plant village dataset with 54,306 images covering 14 crops and 26 crop diseases. Their work was widely appreciated and cited by 943 researchers to date but it had some limitations. The dataset was collected in a controlled environment with a consistent plain background and proper illumination. Deep learning-based models trained on this dataset reports very high accuracies typically 99% and above. Such high disease identification accuracies are phenomenon if they perform analogous when deployed.

There are two challenges in the deployment of smartphone-assisted foliar disease identification. The first challenge is the presence of varying backgrounds and clutter which affect the recognition accuracy. This challenge is partially addressed by incorporating foliar images captured in the field environment. Image augmentation and use of classification ensemble also prove helpful in improving generalization. The second challenge in deployment of smartphone-assisted foliar disease identification models is the poor image quality. This challenge is faced due to the particular nature of the problem as the mobile devices are inherently noisy, used without a mounting device (tripod, etc.) causing loss of structure and blur. Moreover, improper ambient light and amateur photography skills of the user make the image less useful for identification (Barbedo, 2019; Garg *et al.,* 2021).

Machine learning and recently deep learning is increasingly being used in agriculture due to advancement in computing and information processing resources. Deep learning, especially Convolution neural networks (CNN), has shown impressive performance in visual recognition. Computing is used in agriculture in either agricultural data (Ahmed *et al.,* 2021), remote sensing (Abbas *et al.,* 2021; Shafi *et al.,* 2020a,b; Weiss *et al.,* 2020), or camera images. The images are used in different agricultural systems such as leaf-based plant species identification (Ahmed *et al.,* 2016; Saleem *et al.,* 2019), flower classification (Nguyen *et al.,* 2016), weed classification (Olsen *et al.,* 2019) and plant disease classification from either its fruit (Barot and Limbad, 2015), leaf (Hughes and Salathé, 2015) or the mites (Pattnaik *et al.,* 2020). Several works are focusing on leaf-based crop disease identification either focusing on a single crop or multiple crops.

Foliar disease identification is targeted by researchers (Hughes and Salathé, 2015) proposed the largest publicly available dataset of leaf images affected by different plant pathologies. Although, most of the methods have reported impressive accuracies they fail to generalize in field conditions. The datasets are usually collected in a controlled environment and therefore algorithms trained on these datasets perform poorly in field conditions. Some researchers have targeted the problem by using field images (Argüeso *et al.,* 2020; Garg *et al.,* 2021; Ramcharan *et al.,* 2017), whereas the others have incorporated data augmentation (Caluña, *et al.,* 2020; Too *et al.,* 2019). Although these two approaches when combined carefully are enough for most of the task but foliar disease identification, images are supposed to be captured with mobile devices which may face distortions such as motion blur, loss of focus, non-uniform illumination, and high noise which may cause the disease identification system to fail or results in erroneous classification.

We have therefore proposed to use image quality assessment to identify the images of sufficient visual quality before further processing. In this case, the images with visible distortions was discarded at first stage and the classification performance improved. Further, we incorporated an efficient architecture followed by transfer learning and data augmentation to improve the learning and robustness of the trained model which resulted in good generalization as compared to existing approaches.

## MATERIALS AND METHODS

This study was conducted in 2020 at Department of Computer Science and Engineering, University of Engineering and Technology, Lahore, Pakistan. The design of a foliar disease identification system requires a dataset with sufficient number of leaf images to train a classification model. Most of the research in this domain was done by choosing a crop and few of its diseases. Plant village is a public dataset (largest to date) with 54,306 leaf images. The dataset contains 14 crop species: tomato, strawberry, squash, soybean, raspberry, potato, bell pepper, peach, orange, grape, corn, cherry, and apple. The leaf images are affected by fungal, bacterial, and viral diseases as well as mold and mites. Table 1 provides a list of these crops and diseases. Several researchers have tried to train machine learning models for disease identification and remain successful in the identification of a leaf image with impressive accuracies. The problem arises when it is tried to generalize the model by validating the trained model on individually collected leaf images or validating it in real-time. The reason was that the dataset was collected in a controlled environment and when it was tried to validate in an uncontrolled environment, the classification accuracy falls sharply. The details of the PlantVillage dataset are provided in Table 1 which contains crops along with the scientific and common names of diseases affecting the crop, number of leaf images, and type of disease present on the leaf image.

The generalization of the model was tested on an individually collected dataset set of 227 images which was used for cross-evaluation (Ahmed *et al.,* 2021). These images are collected based on the categories which already exist in the PlantVillage dataset. The images are collected from existing datasets, agricultural extension services, and other online sources.

In this work, we have selected two EfficientNet models B0 and B7 for the training of the foliar disease identification module. The EfficientNet-B0 was the mobile version that was intended for offline identification of disease by using local computational resources whereas the EfficientNet-B7 was intended for online identification through a server that may require internet access. The second architecture was expected to provide more accurate classification but at an expense of computational resources.

## Image quality assessment (IQA)

No-reference image quality assessment was a tough task because of the unavailability of reference information and a lot of distortion sources which may affect the quality of an image. As demonstrated by our previous work (Ahmed and Asif, 2020) deeper and more complex architectures usually provide better IQA performance. NASNet and InceptionResNet-V2 provided the best IQA performance from 13 popular convolutional nevel network (CNN) models. However, these architectures are computationally expensive and require larger memory. Natural scene statistics-based methods (Ahmed *et al.,* 2021) and hybrid methods (deep and handcrafted features based) was also an option but they were less versatile and only useful in scenarios where a limited type of distortions are present. Ensembling CNN (Ahmed and Asif, 2019) was also a useful approach but results in larger and complex models. We have therefore opted to use a simple but efficient model for image quality assessment. Pre-trained EfficientNet-B0 was opted as a base architecture and modified to convert it into an IQA model. The modified architecture contained CNN base of pretrained Efficient Net-B0 followed by some fully connected and regression layers as shown in Fig. 1.

**Table 1. Crop village database reflecting the number of plant leaves images alongwith diseases**

| Crop name | Disease name | | Type | Number of images |
|---|---|---|---|---|
| | Common name | Scientific name | | |
| Apple | Cedar apple rust | *Gymnosporangium juniperi-virginianae* | Fungi | 276 |
| | Apple scab | *Venturia insequalis* | Fungi | 630 |
| | Apple rot | *Botryosphaeria obtuse* | Fungi | 621 |
| | - | - | Healthy | 1645 |
| Blueberry | - | - | Healthy | 1502 |
| Cherry | Powdery mildew | *Podosphaera* Spp | Fungi | 1052 |
| | - | - | Healthy | 854 |
| Corn | Cercosporin leaf spot | *Cercospora zeaemaydis* | Fungi | 513 |
| | Common rust | *Puccinia sorghi* | Fungi | 1192 |
| | Northern leaf blight | *Exserohilum turcicum* | Fungi | 985 |
| | - | - | Healthy | 1162 |
| Grape | Black rot | *Guignardia bidwellii* | Fungi | 1180 |
| | Esca (Black measles) | *Phaeomoniella spp* | Fungi | 1384 |
| | Leaf blight | *Pseudocercospora vitis* | Fungi | 1076 |
| | - | - | Healthy | 423 |
| Orange | Huanglongbing | *Candidatus liberibacter* | Bacteria | 5507 |
| Peach | Bacterial sport | *Xanthomonas campestris* | Bacteria | 2291 |
| | - | - | Healthy | 360 |
| Bell pepper | Bacterial spot | *Xanthomonas campestris* | Bacteria | 997 |
| | None | | Healthy | 1148 |
| Potato | Early blight | *Alternaria solani* | Fungi | 1000 |
| | Late blight | *Phytophthora infestans* | Mold | 1000 |
| | - | - | Healthy | 152 |
| Raspberry | - | - | Healthy | 371 |
| Soybean | - | - | Healthy | 5090 |
| Squash | Powdery mildew | *Sphaerotheca fuliginea* | Fungi | 1835 |
| Strawberry | Leaf scorch | *Diplocarpon earlianum* | Fungi | 1109 |
| | - | - | Healthy | 456 |
| Tomato | Early blight | *Alternaria solani* | Fungi | 1000 |
| | Septoria leaf spot | *Septoria lycopersici* | Fungi | 1771 |
| | Target spot | *Corynespora cassiicola* | Fungi | 1404 |
| | Leaf mold | *Fulvia fulva* | Fungi | 952 |
| | Bacterial spot | *Xanthomonas campestris* | Bacteria | 2127 |
| | Late blight | *Phytophthora infestans* | Mold | 1910 |
| | Yellow leaf curl | *Begomovirus* (Fam. Geminiviridae) | Virus | 5357 |
| | Mosaic virus | Tomato mosaic virus (TOMV) | Virus | 373 |
| | Spider mites | *Tetranychus urticae* | Mite | 1676 |
| | - | - | Healthy | 1592 |

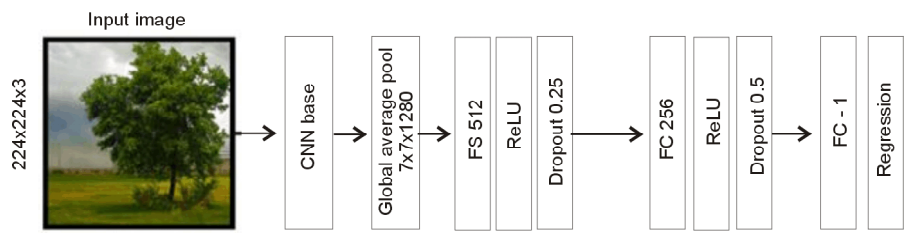

**Fig. 1. CNN architecture for image quality assessment**

The network training was performed using adam optimizer with an initial learning rate of 1e-2 with a learning rate decay of 50% after 15 epochs and trained for a total of 100 epochs. The batch size was set to 32 images with shuffling after each epoch. The model was validated after each epoch and validation patience was set to 10 so the learning was stoped if the validation did not improve for 10 consecutive epochs to avoid over fitting.

## Quality assessment performance

Although the trained model was validated with 3000 images separated from the combined dataset. The model generalization was also validated on two other benchmark databases and the results are given in Table 2. RMSE was a common error measure of a regression model whereas the pearson (PLCC) and spearman's (SROCC) correlation coefficients measure the linear correlation of the predicted quality scores with human subjective scores and the monotonicity of the predictions respectively. The results was very encouraging and indicate good generalization which was in line with the state-of-the-art models of image quality assessment.

**Table 2. Performance of the image quality assessment module on benchmark databases**

|  | BIQ2020+ KonIQ-1K | Live in the Wild CD | CID2013 |
|---|---|---|---|
| RMSE | 0.1482 | 0.1224 | 0.1428 |
| PLCC | 0.8818 | 0.8448 | 0.8512 |
| SROCC | 0.8779 | 0.8424 | 0.8411 |

\* RMSC = Root mean square error, PLCC = Pearson linear correlation coefficient, SROCC = Spearman's rank ordered correaltion coefficient

## Foliar disease identification

Deep learning played a key role in visual recognition and foliar disease identification. CNN is a primary network with an impressive performance in visual recognition. Saleem *et al.* (2020) has provided a list of pre-trained deep architectures, number of parameters, time and epochs required for training, and various performance parameters. Table 3 provides a list of these architectures and performance parameters. Among the models under test, Xception has provided the best validation F1-score which is attributed to its depth wise superable convolution architectural design which is an improvement over conventional inception architecture.

## CNN architecture

The PlantVillage dataset contains two types of leaves: (i) healthy and (ii) unhealthy (Ahmed *et al.,* 2016; Saleem *et al.,* 2019). The architecture used for foliar disease identification is given in Fig. 2. It is notable that global average-pooling layer was replaced with global max-pooling layer followed by fully connected, probability and classification layers. The max-pooling was used due to its ability to discriminate fine-details (Guo *et al.,* 2019).

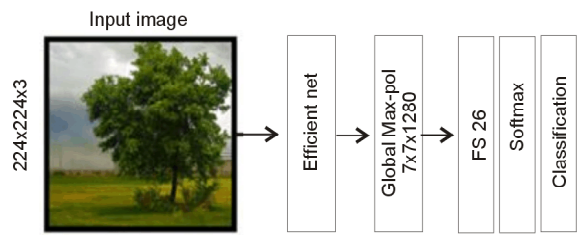

**Fig. 2. Foliar disease identification architecture**

## Preprocessing

Preprocessing was a basic and mandatory step in most computer vision applications. In case of foliar disease identification we performed histogram equalization on the intensity layer and performed the color balance to remove the color cast and adjust the contrast of all the images in this dataset and new images to be identified (Mohanty *et al.,* 2016). We opted to skip the background removal as well, as it worked on this dataset as the background was almost the same but the new test images, varying background and the same background removal method was not worked in that case until specifically trained.

**Table 3. Performance of pre-trained CNN models on plan village dataset along with their size, required number of epochs and training time**

| Architectures | Parameters (millions) | Epochs | Training time (hours) | Training accuracy | Validation accuracy | Training loss | Validation loss | Precision | Recall | F1-scroe |
|---|---|---|---|---|---|---|---|---|---|---|
| LeafNet | 0.324 | 59 | 5.95 | 0.8590 | 0.7961 | 0.4563 | 0.6658 | 0.7946 | 0.7971 | 0.7958 |
| VGG-16 | 138 | 59 | 38.13 | 0.8339 | 0.8189 | 0.5328 | 0.5651 | 0.8182 | 0.8194 | 0.8188 |
| OverFeat | 141.8 | 58 | 6.75 | 0.8995 | 0.8603 | 0.3201 | 0.4330 | 0.8592 | 0.8628 | 0.8610 |
| Improved Cifar-10 | 2.43 | 58 | 6.08 | 0.9256 | 0.8974 | 0.2628 | 0.3205 | 0.8944 | 0.8960 | 0.8952 |
| Inception ResNet-V2 | 54.3 | 58 | 32.83 | 0.9551 | 0.9091 | 0.1530 | 0.3047 | 0.9075 | 0.9105 | 0.9089 |
| Reduced Mobile Net | 0.5 | 55 | 11.72 | 0.9570 | 0.9278 | 0.1860 | 0.2442 | 0.9269 | 0.9267 | 0.9268 |
| Modified Mobile Net | 0.5 | 53 | 6.38 | 0.9534 | 0.9297 | 0.1632 | 0.2385 | 0.9278 | 0.9265 | 0.9271 |
| ResNet-50 | 23.6 | 55 | 26.33 | 0.9873 | 0.9423 | 0.0468 | 0.1923 | 0.9351 | 0.9358 | 0.9354 |
| MLCNN | 78 | 57 | 67.33 | 0.9583 | 0.9402 | 0.1335 | 0.1820 | 0.9386 | 0.9411 | 0.9398 |
| Inception-v4 | 41.2 | 59 | 52.92 | 0.9586 | 0.9489 | 0.1410 | 0.1828 | 0.9410 | 0.9466 | 0.9438 |
| Improved GoogLeNet | 6.8 | 53 | 9.67 | 0.9829 | 0.9521 | 0.0522 | 0.1038 | 0.9528 | 0.9539 | 0.9533 |
| AlexNet | 60 | 54 | 6.10 | 0.9689 | 0.9578 | 0.1046 | 0.1298 | 0.9563 | 0.9570 | 0.9566 |
| DenseNet-121 | 7.1 | 56 | 28.75 | 0.9826 | 0.9580 | 0.0758 | 0.1323 | 0.9581 | 0.9569 | 0.9575 |
| MobileNet | 3.2 | 47 | 14.70 | 0.9764 | 0.9632 | 0.0903 | 0.1090 | 0.9624 | 0.9612 | 0.9618 |
| AgroAVNET* | 238 | 54 | 49.90 | 0.9841 | 0.9649 | 0.0546 | 0.1078 | 0.9626 | 0.9674 | 0.9650 |
| ZFNet | 58.5 | 47 | 6.47 | 0.9752 | 0.9717 | 0.0746 | 0.1139 | 0.9746 | 0.9751 | 0.9748 |
| AlexNet+GoogleNet | 5.6 | 57 | 6.5 | 0.9931 | 0.9818 | 0.0229 | 0.0592 | 0.9749 | 0.9751 | 0.9750 |
| Xception | 22.8 | 34 | 56.28 | 0.9990 | 0.9798 | 0.0140 | 0.0621 | 0.9764 | 0.9767 | 0.9765 |

*AgroAVNET : Hybrid AlexNet with VGG

**Augmentation strategies**

CNNs are inherently robust to orientation, scale and translation but if all the images are presented in the same configuration. To make the model more robust to these variations, we introduced some random variations which served two purposes. The principal purpose was to introduce variations in the data to make it robust. The second objective was to increase the dataset size by repeating the same images after variations in certain aspects. The list of these augmentation strategies along with their effective ranges is givem in Table 4.

Table 4 Augmentation strategies and ranges

| Sr. No. | Augmentation type | Range |
|---------|-------------------|-------|
| 1 | Horizontal reflection | [1, 0] |
| 2 | Vertical reflection | [1,0] |
| 3 | Rotation | [-45,45] |
| 4 | Scaling | [0.75, 1.25] |
| 5 | Translation | [-10, 10] |

Horizontal and vertical reflections ensured that the model did not fix to certation orientation of the leaf image and results in flipped augmented images. Rotation was randomly applied up to $45^O$ in clockwise or anticlockwise direction. Scaling result in randomly applied scaled images in horizontal or vertical direction, as well as compound scaled images. Translation resulted in horizontal or vertically translated images. Moreover, some augmentation strategies which were commonly used in literature such as shear, jitter, contrast, or color change were not applied as they may vary the leaf shape, pattern, or artifacts and will confuse the classifier. Moreover, Fig. 3 elaborates a sample image and the result of some augmentation strategies applied to it.

**Training procedures**

The network training was performed on Dell Precision 5600 workstation with two Intel Xeon-2687 processors, 36GB of RAM, a SATA SSD for dataset storage, and GTX 1070Ti graphics processing unit with 8 GB of GDDR6 memory. The learning optimizer was Adam with an initial learning rate of 3e-3 with a learning rate decay of 50% after 20 epochs. Different batch sizes were used for both models and set to the maximum batch size which was 32 for EfficientNet-B0 and 16 for EfficientNet-B7.The training was performed until there was no improvement in validation accuracy for 3 consecutive epochs. The trained model was validated with a 20% holdout dataset and then the final model was cross-validated for generalization on a self-collected dataset of 224 images as stated in our work. Fig. 4 provided the workflow image quality assessment based foliar disease identification.

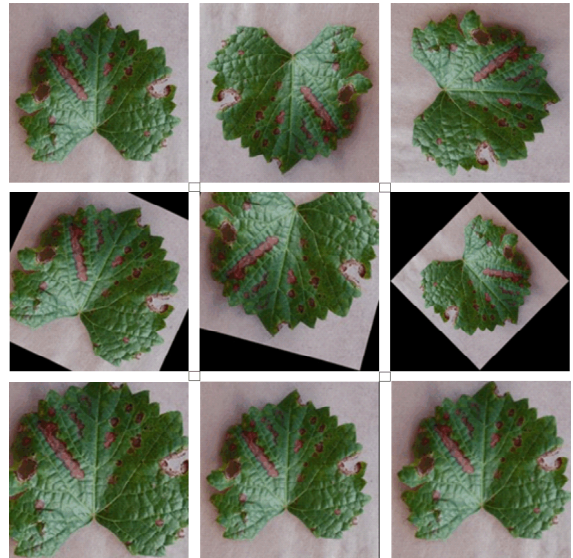

**Fig. 3. Some sample images generated using image augmentation**

**RESULTS AND DISCUSSION**

Objective Image quality assessment is a helpful tool in the automated assessment of the quality of digital images. The remaining 186 images of acceptable quality were provided to the classification module and the results were reported (Table 5).

The performance of the proposed model AgroPath (Agricultural Pathologist) was tested using 10-fold cross-validation. Most of the work based on plant village dataset had reported held-out results with 80-20 split or 60-40 split.

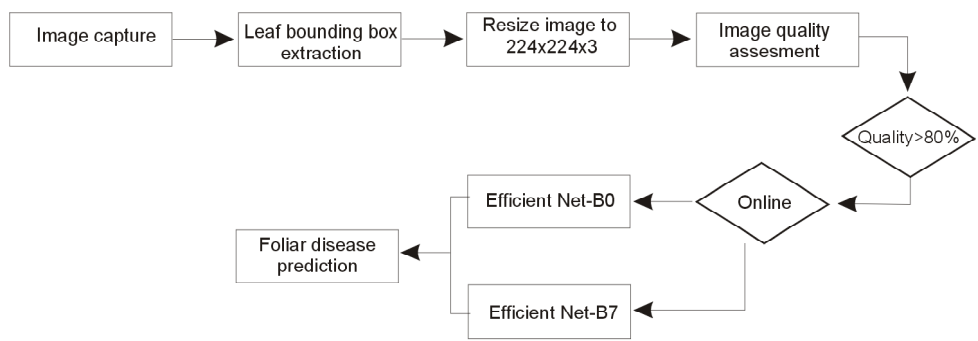

**Fig. 4. Foliar disease prediction workflow**

**Table 5. Results of the agro path approach on plant village**

|  | Precision | Recall | Accuracy (%) | F1-score |
|---|---|---|---|---|
| **Plant village (26 classes of diseases only)** | | | | |
| Ahmed *et al.* (2021) | 99.35 | 99.31 | 99.31 | 99.33 |
| AgroPath-mobile | 99.13 | 99.27 | 99.27 | 99.20 |
| AgroPath-large | 99.76 | 99.79 | 99.79 | 99.77 |
| **Plant village (complete dataset, 38 classes)** | | | | |
| Ahmed *et al.* (2021) | 99.13 | 98.49 | 98.79 | 98.81 |
| AgroPath-mobile | 99.22 | 99.17 | 99.23 | 99.19 |
| AgroPath-large | 99.51 | 99.55 | 99.55 | 99.53 |
| **Self-collected dataset (227 images)** | | | | |
| Ahmed *et al.* (2021) | 82.47 | 82.47 | 82.47 | 82.47 |
| AgroPath-mobile | 99.27 | 99.42 | 99.42 | 99.34 |
| AgroPath-large | 98.38 | 98.38 | 98.38 | 98.38 |

Note: In both model 227 images were used and AgroPath discorded 41 images which was less than Ahmad *et al.* (2021) model.

This hold-out validation approach were easier as only one-time training of the CNN model was required but less reliable as it may be biased due to a certain split of data (Hlaing and Zaw, 2017a). Ahmed *et al.* (2021) reported 10-fold cross-validation accuracy for the complete dataset. In this work, we followed the same cross-validation approach as Ahmed *et al.* (2021) did. Although, took a longer time to train and validate the model 10 times in each iteration the model was trained on 90% data and tested on 10% data which provided more training data with rigorous testing. However, Table 5 provides the results of model training and validation in three different scenarios which were also reported in Ahmed *et al.* (2021). In the first scenario, 26 categories of diseased leaf images were considered for training and validation and the results were compared with Ahmed *et al.* (2021). In the second scenario, the complete dataset with 38 categories with leaf images with or without diseased leaves were used for training and validation of the model. In the third scenario, the trained model in the first scenario along with the quality assessment module was used for validation. It was to be noted, that Ahmed *et al.* (2021) reported results for 227 images whereas our model discarded 41 images due to quality concerns, and only 186 images were used for model validation yielding higher accuracy (99.42) and other performance parameters. This indicated the utility of AgroPath to discard images of low quality and was useful in scenarios where images were proned to severe degradations. The comparison of the AgroPath was made with the existing approaches and provided in Table 6.

**Table 6. Comparison of the AgroPath with existing approaches**

| Existing work (Author and year) | Accuracy (%) | Precision | Recall | F1-Score | Remarks |
|---|---|---|---|---|---|
| (Mohanty *et al.*, 2016) | 99.35 | 99.35 | 99.35 | 99.34 | 38 classes with 60/40 split |
| (Islam *et al.*, 2017) | 95 | 95 | 95 | 95 | 3 classes with 60/40 split |
| (Hlaing and Zaw, 2017b) | 78.7 | - | - | - | 38 classes with 5-fold CV |
| (Hlaing and Zaw, 2017a) | 84.7 | - | - | - | 7 classes with 10-fold CV |
| (Amara *et al.*, 2017) | *99.72* | *99.70* | *99.72* | *99.71* | 3 classes with 50/50 split |
| (Durmuş *et al.*, 2017) | 95.65 | - | - | - | 10 classes with 80/20 split |
| (Brahimi *et al.*, 2017) | *99.185* | *98.529* | *98.532* | *98.518* | 9 classes with 5-fold CV |
| (Yamamoto *et al.*, 2017) | 92 | 90 | 92 | 91 | 9 classes with 80/20 split |
| (Yuan *et al.*, 2018) | 95.97 | - | - | - | 38 classes with 80/20 split |
| (Goncharov *et al.*, 2018) | *99* | - | - | - | 5 classes with 75/25 split |
| (Wang *et al.*, 2018) | 90.84 | - | - | - | 8 classes with 80/20 split |
| (Zeng *et al.*, 2018) | 91.79 | - | - | - | 38 classes with 80/20 split |
| (Ferentinos, 2018) | 99.53 | - | - | - | 58 classes with 80/20 split |
| (Yadav *et al.*, 2018) | 97.39 | 97.76 | 97.39 | 97.34 | 23 classes with 70/30 split |
| (Kour and Arora, 2019) | 95.23 | - | - | - | 7 classes with 70/15/15 split |
| (Pardede *et al.*, 2018) | 87.01 | - | - | - | 7 classes with 80/20 split |
| (Suryawati *et al.*, 2018) | 95.24 | - | - | - | 10 classes with 80/10/10 split |
| (Too *et al.*, 2019) | 99.75 | - | - | - | 38 classes with 80/20 split |
| (Khandelwal and Raman, 2019) | 99.37 | - | - | - | 57 classes with 80/20 split |
| (Barbedo, 2019) | 94 | - | - | - | 14 classes with 80/20 split |
| (Baranwal *et al.*, 2019) | 98.54 | - | - | - | 38 classes with 80/20 split |
| (Arsenovic *et al.*, 2019) | 93.67 | - | - | - | 38 classes with 80/20 split |
| (Angin *et al.*, 2020) | 0.9567 | - | - | - | 38 classes with 80/20 split |
| (Saleem *et al.*, 2020) | 0.9981 | 0.9981 | 0.9975 | 0.9978 | 38 classes with 70/15/15 split |
| (Argüeso *et al.*, 2020) | 91.4 | 93.8 | 92.6 | - | 32 classes with 80/20 split (20x) |
| (Agarwal *et al.*, 2020) | 98.4 | - | - | - | 10 classes with 80/20 split |
| (Chen *et al.*, 2021) | 99.12 | 99.01 | 99.94 | - | 38 classes with 80/20 split |
| (Verma *et al.*, 2020) | 91.83 | - | - | - | 38 classes with 80/20 split |
| (Caluña *et al.*, 2020) | 93 | 91 | 97 | 94 | 38 classes with 80/20 split |
| (Chen *et al.*, 2021) | 93.75 | - | - | - | 38 classes with 80/20 split |
| (Bansal and Kumar, 2021) | 96.05 | - | - | - | 4 classes with 80/20 split |
| (Leelavathy and Kovvur, 2021) | 98.33 | - | - | - | 38 classes with 80/20 split |
| (Malik *et al.*, 2021) | 93.40 | - | - | - | 4 classes with 80/20 split |
| (Ahmed *et al.*, 2021) | 99.35 | 99.31 | 99.31 | 99.33 | 26 classes with 10-fold CV |
| Agro Path-mobile | 99.13 | 99.27 | 99.27 | 99.20 | 38 classes with 10-fold CV |
| Agro Path-large | 99.76 | 99.79 | 99.79 | 99.77 | 38 classes with 10-fold CV |
| Agro Path-mobile | 99.22 | 99.17 | 99.23 | 99.19 | 26 classes with 10-fold CV |
| Agro Path-large | 99.51 | 99.55 | 99.55 | 99.53 | 26 classes with 10-fold CV |

Although Ferentinos (2018) and Khandelwal and Raman (2019) reported performance on 58 and 57 classes the dataset with this number of classes was not publicly available. Both of these works reported 99.53% and 99.37% accuracies respectively with an 80/20 train/test split. Fourteen researchers have trained and validated their model on 38 classes with most of them reporting accuracy with an 80/20 split. Mohanty *et al.* (2016) reported the performance with a 60/40 split, Hlaing and Zaw (2017b) reported 5-fold cross-validation performance and Saleem *et al.* (2020) reported performance with 70/15/15 split with train/validation/test. Ahmed *et al.* (2021) has reported the performance with 10-fold cross-validation for 38 classes as well as 26 disease classes. Other works, used a subset of the classes based on either disease or crops. It was to be noted that most of the carefully designed deep learning approaches easily achieve 99%+ accuracies when trained using better training strategies such as data augmentation and pre-processing. The proposed approach was intended for generalizability and we had performed a cross-dataset evaluation on a self-collected dataset of images. The same dataset was used by Ahmed *et al.* (2021) who has presented a texture and color features-based approach. Their approach is intended for a mobile platform without an embedded graphics processing unit and internet access, however, our approach was also targeted for low computational complexity at the user end and higher computational complexity at the server-side.

## CONCLUSION

The study demonstrates the viability of smartphone assisted foliar disease identification. Contrary to the existing studies, we have targeted the generalization of the trained model for the practical and noisy environment and introduced the image quality assessment module to perform initial screening before performing disease identification. We have proposed two architectures of AgroPath one is mobile architecture for deployment on smartphone and the other is a larger architecture which can be used for server end processing with a more accurate diagnosis. Both the architectures have performed better compared to the state-of-the-art using the benchmark dataset PlantVillage and the demonstrated performance on a self-collected dataset was also reasonable. The recommendations are based on deep learning models for smartphone-based identification of weeds and plant pathologies as they may prove helpful in early diagnosis and prevention of diseases and avoid further crop loss.

## CONTRIBUTION OF AUTHORS

| Sr. No. | Author's name | Contribution | Signature |
|---------|---------------|--------------|-----------|
| 1 | Nisar Ahmed | Conducted the research work and wrote the manuscript | 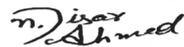 |
| 2 | Hafiz Muhammad Shahzad Asif | Supervised the research work | 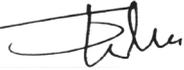 |
| 3 | Gulshan Saleem | Helped in data analysis | 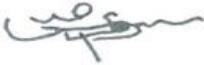 |
| 4 | Muhammad Usman Younus | Helped in the research work and proof read the manuscript | 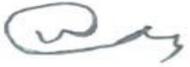 |